%% file: main.tex
\documentclass[10pt,letterpaper]{article}

\usepackage[letterpaper,margin=0pt]{geometry}
\usepackage[T1]{fontenc}
\usepackage{lmodern}
\usepackage{fontspec}
\usepackage{xcolor}
\usepackage{graphicx}
\usepackage{tikz}
\usepackage{etoolbox}
\usepackage{ragged2e}
\usepackage{microtype}
\usepackage{amsmath}
\usepackage{amssymb}
\usepackage{booktabs}
\usepackage{tabularx}
\usepackage{longtable}
\usepackage{enumitem}
\usepackage{titlesec}
\usepackage{fancyhdr}
\usepackage[hidelinks]{hyperref}
\usetikzlibrary{calc,arrows.meta,positioning}

\newfontface\brandfontface{Poppins-ExtraBoldItalic.ttf}[Path=assets/fonts/]

\definecolor{Ink}{HTML}{15243A}
\definecolor{MutedInk}{HTML}{5B6C83}
\definecolor{BrandNavy}{HTML}{00162E}
\definecolor{BrandBlue}{HTML}{0B70B5}
\definecolor{BrandBright}{HTML}{229BD2}
\definecolor{BrandDeepBlue}{HTML}{064A73}
\definecolor{TitleMuted}{HTML}{0A1B36}
\definecolor{FigureEdge}{HTML}{BED8E9}
\definecolor{FigureHint}{HTML}{7794AA}

\newcommand{\coverbackgroundasset}{assets/cover-background.png}
\newsavebox{\coverimagebox}
\newcommand{\coverbackground}{%
  \sbox{\coverimagebox}{\includegraphics[width=\paperwidth]{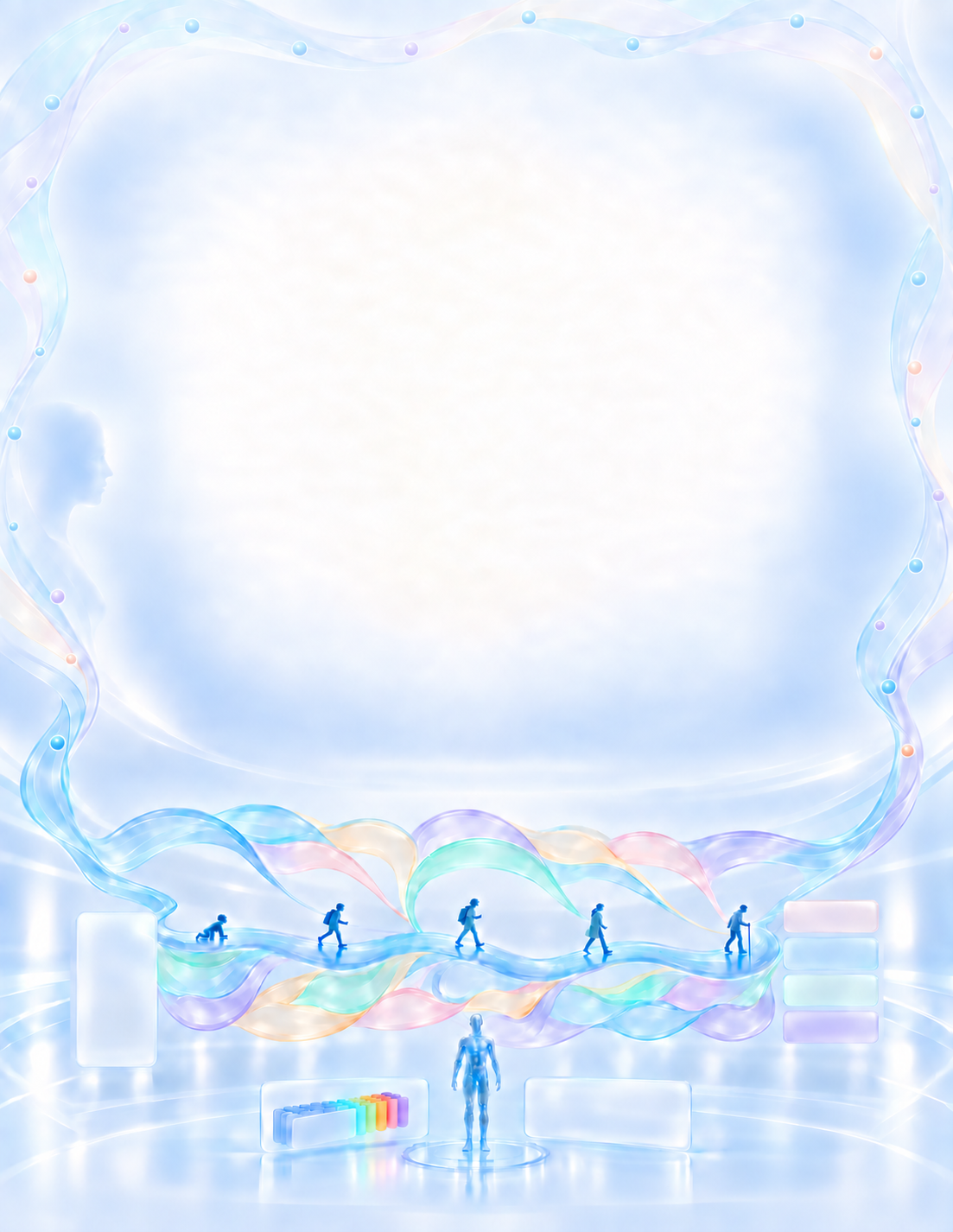}}%
  \begin{tikzpicture}[remember picture,overlay]
    \clip (current page.south west) rectangle (current page.north east);
    \ifdim\ht\coverimagebox<\paperheight
      \node[inner sep=0pt,outer sep=0pt] at (current page.center)
        {\includegraphics[height=\paperheight]{\coverbackgroundasset}};
    \else
      \node[inner sep=0pt,outer sep=0pt] at (current page.center)
        {\usebox{\coverimagebox}};
    \fi
  \end{tikzpicture}%
}

\newcommand{\modelingfigureasset}{assets/graphical-abstract.pdf}
\newcommand{\covermodelingfigure}{%
  \begin{tikzpicture}[remember picture,overlay]
      \coordinate (modelcenter) at ([yshift=2.27in]current page.south);
      \fill[white,fill opacity=0.08,rounded corners=11pt]
        ([xshift=-3.655in,yshift=-2.071in]modelcenter) rectangle
        ([xshift=3.655in,yshift=2.071in]modelcenter);
      \begin{scope}
        \clip[rounded corners=8pt]
          ([xshift=-3.62in,yshift=-2.036in]modelcenter) rectangle
          ([xshift=3.62in,yshift=2.036in]modelcenter);
        \node[inner sep=0pt] at (modelcenter)
          {\includegraphics[width=7.24in,height=4.072in,keepaspectratio]{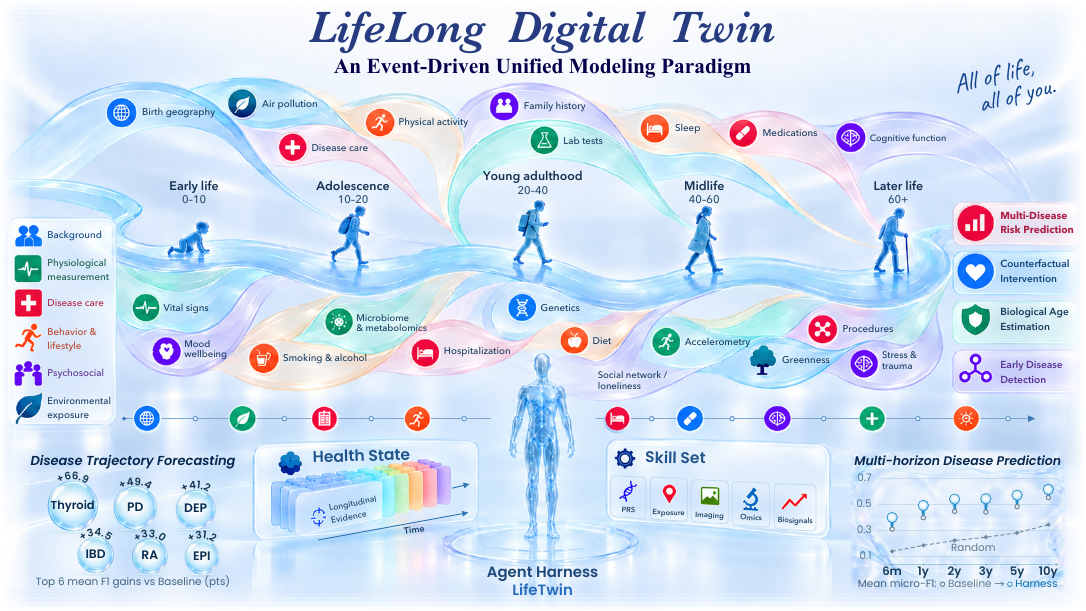}};
      \end{scope}
      \begin{scope}
        \pgfseteorule
        \clip
          ([xshift=-3.80in,yshift=-2.20in]modelcenter) rectangle
          ([xshift=3.80in,yshift=2.20in]modelcenter)
          ([xshift=-3.75in,yshift=0.50in]modelcenter) rectangle
          ([xshift=-3.56in,yshift=1.35in]modelcenter)
          ([xshift=3.56in,yshift=0.75in]modelcenter) rectangle
          ([xshift=3.75in,yshift=1.60in]modelcenter);
        \draw[white,draw opacity=0.50,line width=2.2pt,rounded corners=11pt]
          ([xshift=-3.655in,yshift=-2.071in]modelcenter) rectangle
          ([xshift=3.655in,yshift=2.071in]modelcenter);
        \draw[FigureEdge!75!BrandBlue,draw opacity=0.72,
          line width=0.60pt,rounded corners=11pt]
          ([xshift=-3.655in,yshift=-2.071in]modelcenter) rectangle
          ([xshift=3.655in,yshift=2.071in]modelcenter);
      \end{scope}
      \foreach \rimx/\rimlow/\rimhigh in {-3.655/0.50/1.35,3.655/0.75/1.60} {
        \foreach \band in {0,...,7} {
          \pgfmathsetmacro{\rimalpha}{(7.5-\band)/8}
          \pgfmathsetmacro{\rimwhite}{0.50*\rimalpha}
          \pgfmathsetmacro{\rimblue}{0.72*\rimalpha}
          \pgfmathsetmacro{\rimbottoma}{\rimlow+0.02*\band}
          \pgfmathsetmacro{\rimbottomb}{\rimlow+0.02*(\band+1)}
          \pgfmathsetmacro{\rimtopa}{\rimhigh-0.02*(\band+1)}
          \pgfmathsetmacro{\rimtopb}{\rimhigh-0.02*\band}
          \foreach \rima/\rimb in {
            \rimbottoma/\rimbottomb,\rimtopa/\rimtopb} {
            \draw[white,draw opacity=\rimwhite,line width=2.2pt,line cap=butt]
              ([xshift=\rimx in,yshift=\rima in]modelcenter) --
              ([xshift=\rimx in,yshift=\rimb in]modelcenter);
            \draw[FigureEdge!75!BrandBlue,draw opacity=\rimblue,
              line width=0.60pt,line cap=butt]
              ([xshift=\rimx in,yshift=\rima in]modelcenter) --
              ([xshift=\rimx in,yshift=\rimb in]modelcenter);
          }
        }
      }
  \end{tikzpicture}%
}

\newcommand{\paperbrandbase}{LifeLong Digital}
\newcommand{\paperbrandaccent}{Twin}
\newcommand{\papertitle}{\paperbrandbase\space\paperbrandaccent}
\newcommand{\papersubtitle}{A Unified Modeling Paradigm and Agent Harness}
\newcommand{\paperscope}{for Event-Driven Lifelong Health State Trajectories}

\newcommand{\brandfont}{%
  \brandfontface\fontsize{47}{52}\selectfont%
}
\newcommand{\coveraccentword}{%
  {\color{BrandDeepBlue!65!BrandBlue}T}%
  \kern-0.025em%
  {\color{BrandDeepBlue!30!BrandBlue}w}%
  {\color{BrandBlue}i}%
  {\color{BrandBlue!55!BrandBright}n}%
}
\newcommand{\coverwordmark}{%
  \resizebox{4.85in}{!}{%
    {\brandfont\color{BrandNavy}\paperbrandbase\space}%
    {\brandfont\coveraccentword}%
  }%
}

\newcommand{\covermasthead}{%
  \makebox[\linewidth][c]{\coverwordmark}%
}

\newcommand{\coversubtitlestyle}{%
  \sffamily\mdseries\itshape\fontsize{18}{23}\selectfont\color{TitleMuted}%
}
\newcommand{\titlekeyword}[1]{\textcolor{BrandBlue}{\textbf{#1}}}
\newcommand{\coversubtitle}{A Unified \titlekeyword{Modeling Paradigm} and Agent \titlekeyword{Harness}}

\newcommand{\mailicon}{%
  \tikz[baseline=-0.34ex,x=1.20ex,y=1.20ex]{%
    \draw[draw=BrandBlue,line width=0.45pt,rounded corners=0.07ex]
      (0,0) rectangle (1,0.68);
    \draw[draw=BrandBlue,line width=0.45pt,line cap=round,line join=round]
      (0.05,0.61) -- (0.50,0.30) -- (0.95,0.61);
  }%
}
\newcommand{\paperauthors}{%
  \begin{tabular}{@{}c@{\hspace{17pt}}c@{\hspace{17pt}}c@{\hspace{17pt}}c@{}}
    Jin Jiang\textsuperscript{1,2} &
    Sean Yates\textsuperscript{1,2} &
    Jasper Chong\textsuperscript{2} &
    Raymond Brooks\textsuperscript{2} \\[4pt]
    \multicolumn{4}{c}{%
      Alex Lawson\textsuperscript{2}\hspace{28pt}%
      Yuqin Qin\textsuperscript{1}\hspace{28pt}%
      Liangcai Gao\textsuperscript{\mailicon\hspace{0.15em}1}%
    }
  \end{tabular}%
}
\newcommand{\paperaffiliations}{%
  \textsuperscript{1}Peking University\qquad
  \textsuperscript{2}VITA-Med Group\par
}
\newcommand{\papercorrespondence}{%
  Correspondence:
  \href{mailto:jiangjin@stu.pku.edu.cn}{jiangjin@stu.pku.edu.cn}%
  ,
  \href{mailto:gaoliangcai@pku.edu.cn}{gaoliangcai@pku.edu.cn}%
}
\hypersetup{
  pdftitle={\papertitle: \papersubtitle\space\paperscope},
  pdfauthor={Jin Jiang, Sean Yates, Jasper Chong, Raymond Brooks, Alex Lawson, Yuqin Qin, Liangcai Gao},
  pdfsubject={\papersubtitle\space\paperscope}
}

\newcommand{\coverabstract}{%
Human health is a continuous, dynamic trajectory shaped by the cumulative
interplay of biological processes, clinical events, behaviors and environmental
exposures across the life course. Unifying the full breadth of lifelong health
information, including longitudinal records, genetic variation, molecular
profiles and environmental histories, is essential for whole-person modeling
and remains a major challenge. We introduce LifeLong Digital Twin, a unified,
event-driven modeling paradigm that organizes Life Events into daily Health
States and accumulates them into Lifelong Health Context. An accompanying Agent
Harness incorporates multimodal evidence beyond the language model's textual
context. We evaluate four language models across
25 disease endpoints on three tasks: Disease \mbox{Trajectory} Forecasting, Disease
Risk Ranking and Multi-horizon Disease Prediction. The approach yields marked gains over the reference condition:
model-averaged F1 increases by 22.0\% for disease identification in
trajectory forecasting and 18.3\% for five-year disease outcomes; thyroid-disease
F1 reaches 0.669. The framework provides a foundation for
whole-person digital twins and research on personalized lifelong disease
prevention.
}

\fancypagestyle{manuscript}{%
  \fancyhf{}
  \fancyhead[L]{\sffamily\fontsize{8}{10}\selectfont\color{MutedInk}LifeLong Digital Twin}
  \fancyhead[R]{}
  \fancyfoot[C]{\sffamily\fontsize{9}{11}\selectfont\color{MutedInk}\thepage}

}

\begin{document}
\color{Ink}

\AddToHookNext{shipout/background}{\coverbackground}
\AddToHookNext{shipout/foreground}{\covermodelingfigure}
\vspace*{0.78in}
\noindent\hspace*{1.00in}%
\begin{minipage}[t]{6.50in}
  \centering
  \covermasthead\par

  \vspace{18pt}
  \makebox[\linewidth][c]{%
    \begin{minipage}{7.20in}
      \centering
      \coversubtitlestyle
      \mbox{\coversubtitle}\par
      \vspace{5pt}
      \mbox{\paperscope}\par
    \end{minipage}%
  }\par

  \vspace{26pt}
  {\sffamily\bfseries\fontsize{11.2}{14}\selectfont
    \paperauthors\par}

  \vspace{12pt}
  \begin{minipage}{5.80in}
    \centering
    \sffamily\fontsize{9.2}{11.5}\selectfont
    \paperaffiliations
  \end{minipage}\par

  \vspace{10pt}
  {\sffamily\color{MutedInk}\fontsize{8.5}{10.5}\selectfont
    \papercorrespondence\par}

  \vspace{14pt}
  \begin{minipage}{5.90in}
    \justifying
    \setlength{\emergencystretch}{1em}
    \hyphenpenalty=600
    \exhyphenpenalty=100
    \fontsize{10.5}{12.8}\selectfont
    \noindent\textcolor{BrandBlue}{\textbf{Abstract.}}\enspace\coverabstract
  \end{minipage}

\end{minipage}

\clearpage
\newgeometry{left=0.88in,right=0.88in,top=0.83in,bottom=0.83in,
  headheight=14pt,headsep=16pt,footskip=25pt}
\setlength{\headwidth}{\textwidth}
\pagestyle{manuscript}
\rmfamily\mdseries\upshape\fontsize{10.5}{13.5}\selectfont
\color{Ink}
\setlength{\parindent}{0pt}
\setlength{\parskip}{4pt}
\setlength{\emergencystretch}{1.5em}
\setlist[itemize]{leftmargin=1.4em,itemsep=2pt,topsep=4pt,parsep=0pt}
\setlist[enumerate]{leftmargin=1.6em,itemsep=3pt,topsep=4pt,parsep=0pt}
\titleformat{\section}{\sffamily\bfseries\fontsize{15}{19}\selectfont\color{BrandNavy}}{\thesection}{0.65em}{}
\titleformat{\subsection}{\sffamily\bfseries\fontsize{11.5}{15}\selectfont\color{BrandDeepBlue}}{\thesubsection}{0.65em}{}
\titleformat{\subsubsection}{\sffamily\bfseries\fontsize{10.5}{14}\selectfont\color{Ink}}{\thesubsubsection}{0.65em}{}
\titlespacing*{\section}{0pt}{15pt}{8pt}
\titlespacing*{\subsection}{0pt}{11pt}{5pt}
\titlespacing*{\subsubsection}{0pt}{9pt}{3pt}
\setcounter{secnumdepth}{3}
\makeatletter\setlength{\@fptop}{0pt}\makeatother
\input{sections/01-introduction}
\clearpage
\input{sections/02-lifelong-digital-twin}
\clearpage
\input{sections/03-experimental-setup}
\clearpage
\input{sections/04-results}
\input{sections/05-future-work}

\clearpage

\input{references}
\clearpage
\input{sections/06-appendix}

\end{document}

%% file: sections/01-introduction.tex
\section{Introduction}
\label{sec:introduction}

Human health unfolds through a lifetime of interacting biological processes,
clinical encounters, behaviors and environmental exposures. Their effects
accumulate across years, while the evidence available to describe them is
fragmented across settings and moments in time. A diagnosis, an image or a
change in physical activity gains meaning from the history of the person
behind it. Realizing the promise of digital twins for personalized
medicine~\cite{bjornsson2020} therefore requires a coherent, evolving account of the individual.

Existing approaches address selected aspects of personal health modeling.
\mbox{Delphi-2M} represents diagnoses and selected health attributes as discrete
tokens, using longitudinal information for disease prediction. It is a
dedicated GPT-based disease model, rather than an application of a pretrained
general-purpose language model~\cite{shmatko2025}. Work on generalist medical
AI and language-model agents offers complementary directions for multimodal
interpretation and reasoning~\cite{moor2023,yao2023}. Bringing these capabilities
together requires a unifying paradigm that organizes diverse personal
evidence across life stages into a coherent, evolving health context.

We introduce \emph{LifeLong Digital Twin}, a unified, event-driven modeling
paradigm for integrating personal health information across the life course.
Its organizing principle is \emph{Lifelong Health Context}: the accumulated
history through which observations become meaningful in relation to one
another. The framework is designed to bring evidence from clinical
narratives, medical imaging, genetic information, physical-activity
trajectories and environmental exposures into this shared context. It places
the individual and their evolving history at the center of modeling,
connecting information across health domains and life stages. This provides
large language models (LLMs) with a common foundation for interpreting new
observations and anticipating future disease.

The paradigm begins with a unified organization of personal experience as
\emph{Life Events}. Events occurring or observed on the same day form a
\emph{Health State}, and successive states accumulate into Lifelong Health
Context. An \emph{Agent Harness} builds on this history, bringing specialized
expertise to the individual's health questions. It relates clinical,
molecular, physiological and environmental evidence to the broader
personal context, supporting reasoning across sources and over time.
Together, event-driven modeling and the Agent Harness enable an evolving
digital twin in which new evidence enriches the history available for
subsequent health reasoning.

Evaluation spans 25 disease endpoints and three complementary
tasks: Disease Trajectory Forecasting, Disease Risk Ranking and
Multi-horizon Disease Prediction. Compared with the baseline setting, the
full framework delivers broad improvements in disease-event identification,
diagnosis-year agreement and prediction across short and long horizons.
Particularly large identification gains for thyroid disease and Parkinson
disease illustrate the potential of the approach across distinct clinical
domains. These findings support the predictive value of Lifelong Health
Context and establish a foundation for whole-person digital twins and
research on personalized lifelong disease prevention.

%% file: sections/02-lifelong-digital-twin.tex
\section{LifeLong Digital Twin}
\label{sec:method}

\subsection{A Unified Event-Driven Modeling Paradigm}
\label{sec:paradigm}

\begin{figure}[!htbp]
 \centering
 \includegraphics[width=\linewidth]{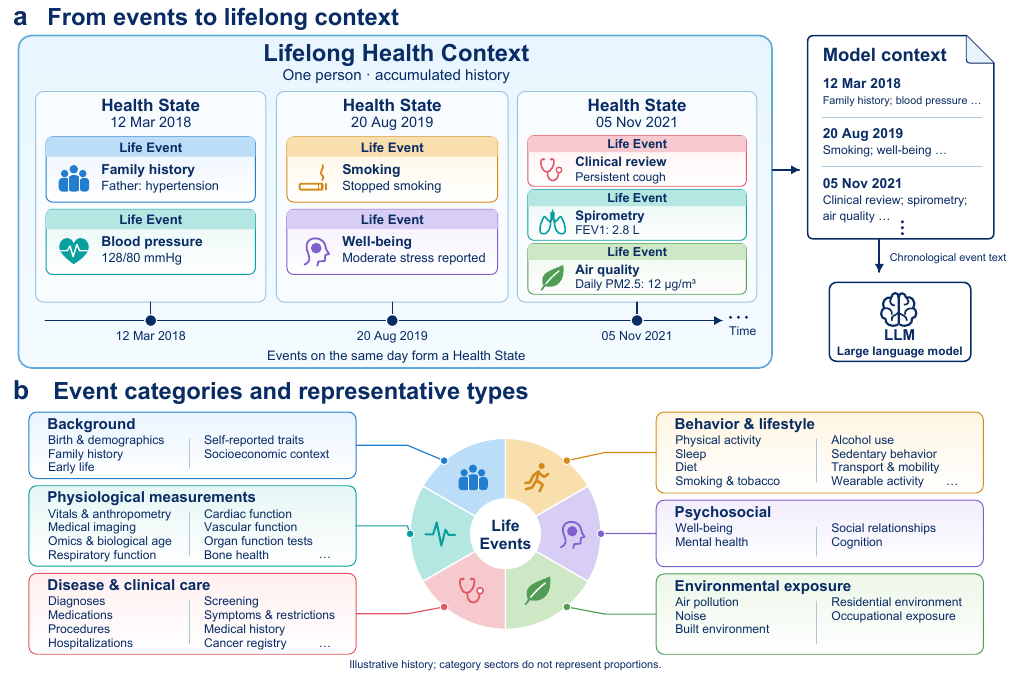}
 \caption{From Life Events to Lifelong Health Context. \textbf{a}, Seven
 synthetic Life Events are grouped into three daily Health States containing
 two, two, and three events. The ordered states form an accumulated history,
 expressed as chronological event text for an LLM. Earlier events remain in
 the history without being copied into later states; ellipses indicate
 continuation. \textbf{b}, Six event categories and representative finer
 types. Colors link events to categories; the equal-sized sectors organize
 the taxonomy and do not indicate data proportions. The same illustrative
 history is expanded in Figure~\ref{fig:context-harness}a.}
 \label{fig:modeling-structure}
\end{figure}

LifeLong Digital Twin organizes a person's evolving history into individual
\emph{Life Events}, daily \emph{Health States}, and the accumulated
\emph{Lifelong Health Context}. This organization connects observations across
health domains and life stages, providing a shared basis for interpreting
new evidence and anticipating future disease. Figure~\ref{fig:modeling-structure}
illustrates this organization through a synthetic history and a taxonomy
of event categories.

\subsubsection{Life Events}

A Life Event is an information unit describing a personal experience,
observation, or change. We write it as $e_i=(\tau_i,c_i,x_i)$, where $\tau_i$
is its occurrence or observation time, $c_i$ its category and type, and $x_i$
its content. Source references, units, modality, recording time, and evidence
availability accompany the event where applicable. A family-history report,
for example, is an observation made on the report date and can also refer to
an earlier period in a relative's life.

Six categories organize the events: \textbf{Background} (demographics, family
history, early life); \textbf{Physiological measurements} (vital signs,
laboratory and functional tests); \textbf{Disease and clinical care}
(diagnoses, medications, procedures); \textbf{Behavior and lifestyle}
(diet, activity, smoking, sleep); \textbf{Psychosocial} (well-being,
relationships, stressors); and \textbf{Environmental exposure} (air pollution,
noise, neighborhood conditions). Finer event types describe the observation
within each category. Categories classify the available events; they do not
specify fields that every daily state must contain.

The event category describes what the information concerns, while modality
describes its form. Text, codes, measurements, images, molecular assays, and
signals can support different event categories. The source observation,
a derived measurement, and an interpretation retain their respective
provenance and evidential roles.

\subsubsection{Daily Health States and Lifelong Health Context}
\label{sec:formulation}

For person $p$, a Health State on day $d$ is the set of events occurring or
observed on that day:
\begin{equation}
 S_p(d)=\{e_i:\operatorname{person}(e_i)=p,\ \tau_i\text{ resolves to day }d\},
 \qquad S_p(d)\ne\varnothing .
 \label{eq:health-state}
\end{equation}
A state contains only the events of its own day. Days without events create
no additional state. Thus a family-history observation and a blood-pressure
measurement can form one state, while later reports of stopping smoking
and moderate stress form another. The earlier observations remain in the history; the later
state does not restate them or infer an updated blood pressure.

For day-resolved events, the accumulated history through day $t$ is
\begin{equation}
 \mathcal{C}_p(t)=\langle S_p(d_1),S_p(d_2),\ldots,S_p(d_m)\rangle,
 \qquad d_1<d_2<\cdots<d_m\leq t .
 \label{eq:context}
\end{equation}
This ordered component, together with events retaining coarser or unresolved
time information, constitutes \emph{Lifelong Health Context}. Month-only,
year-only, and interval-dated events are carried alongside the daily states
at their original precision. They are not assigned an invented day or forced
into a same-day group. Recording time is stored separately from occurrence
or observation time, so a later entry can add information about an earlier
period while preserving when it was recorded.

Health State is an organization of recorded events, and Lifelong Health
Context is the cumulative record available for longitudinal reasoning.
Neither requires complete observation of a person's life. Evidence may be
sparse, missing, or conflicting, and health can change between observations.
The prediction tasks and their evaluation settings are defined in
Section~\ref{sec:experiments}.

\subsection{Direct Reasoning with Lifelong Health Context}
\label{sec:context}

Lifelong Health Context provides a common information basis for personal
health reasoning. In the direct \emph{Context} formulation, the accumulated
history is expressed as chronological event text and supplied to an LLM
together with the task. The text preserves the grouping of events into
Health States and the sequence through which the history develops
(Figure~\ref{fig:context-harness}a). The model therefore receives both
earlier background and subsequent observations when interpreting a health
question.

This formulation allows a recent measurement, symptom, or lifestyle change
to be considered in relation to the individual's prior history. The same
context can support questions about current observations and forecasts of
future disease; the task specifies the requested outcome. This direct use of the accumulated history forms the starting point for
the Agent Harness described next. The task summaries in
Section~\ref{sec:tasks} illustrate the prediction questions
posed against this personal history.

\subsection{An Agent Harness for Personal Health}
\label{sec:harness}

\begin{figure}[!htbp]
 \centering
 \includegraphics[width=\linewidth]{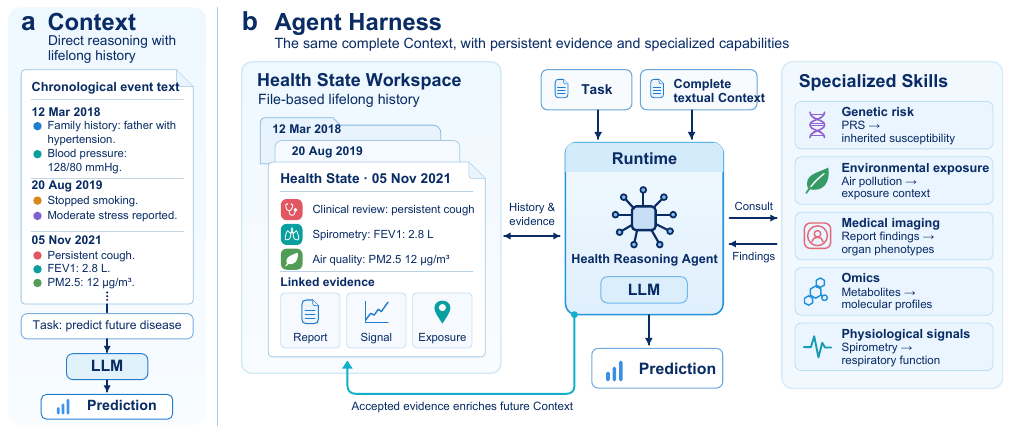}
 \caption{Direct reasoning with Lifelong Health Context and the Agent Harness.
 \textbf{a}, Chronological event text preserves dated Health States and is
 supplied with a task to an LLM. \textbf{b}, The Agent Harness retains the
 same complete textual Context while connecting a file-based Health State
 Workspace, a runtime hosting the Health Reasoning Agent, and Specialized
 Skills. Accepted evidence can enrich subsequent Context; predictions remain
 separate task outputs. The example is synthetic. File cards illustrate
 logical organization rather than a prescribed file per day; Skills show
 representative capabilities.}
 \label{fig:context-harness}
\end{figure}

The \emph{Agent Harness} extends direct reasoning with Context by connecting
the same personal history to persistent evidence and specialized health
capabilities. It retains the complete admitted textual Context as its
foundation and adds three complementary components
(Figure~\ref{fig:context-harness}b): a \emph{Health State Workspace}
that organizes the history and linked evidence, \emph{Specialized Skills}
that relate domain-specific observations to their health meaning, and a
\emph{Health Reasoning Agent} that coordinates these resources around a
task. Together, they support reasoning across sources while allowing
accepted evidence to enrich Lifelong Health Context for later questions.
We refer to this formulation as \emph{Harness} in the evaluation.

\subsubsection{Health State Workspace}
\label{sec:workspace}

The workspace maintains a \emph{file-based Lifelong Health Context} that
connects dated Health States and their events to supporting records and
modality-specific evidence. It provides persistent access to the history
and its details while the agent retains the complete textual Context as
its starting information. Chronological and topic-based views offer
complementary ways to navigate this common record.

New observations enter the history with their own time and provenance.
Accepted derived evidence can also enrich the context, with its derivation
and availability recorded. Source observations, derived evidence, and model
interpretations retain distinct roles. A prediction is a task output and is
not automatically inserted as an event that has occurred. Reuse in later
tasks remains subject to the applicable evidence policy.

\begin{samepage}
\subsubsection{Specialized Skills}
\label{sec:skills}

Specialized Skills connect available domain evidence to its health meaning.
The framework includes five complementary classes:
\begin{itemize}
 \item \textbf{Genetic risk}: an available polygenic risk score (PRS)
 characterizes inherited susceptibility, interpreted with its trait and
 population scope~\cite{lambert2021}.
 \item \textbf{Environmental exposure}: a time-compatible air-pollution
 estimate supplies exposure context for the person's history.
 \item \textbf{Medical imaging}: findings in an imaging report describe
 organ phenotypes associated with the acquisition.
 \item \textbf{Omics}: metabolite measurements characterize molecular
 profiles, with their sample timing and assay context.
 \item \textbf{Physiological signals}: spirometry curves provide evidence
 about respiratory function, with recording and quality information.
\end{itemize}
These examples describe the framework's scope. The task and available
evidence determine which Skills are used; their run-specific configuration
is a separate experimental property. Interpreting an existing imaging
report or PRS uses derived evidence and does not imply analysis of raw
images or genotypes.\par
\end{samepage}

\subsubsection{Health Reasoning Agent}
\label{sec:runtime}

The agent receives the task and complete textual Context, consults the
workspace and applicable Skills, and integrates relevant evidence into its
answer. It relates specialized findings to the individual's broader
history, connecting earlier background, recent events, and complementary
measurements. The Harness thus extends the evidence and capabilities
available around the same Context foundation. Accepted evidence can expand
the record; a forecast remains a possible future outcome, separate from
events that have occurred. The reported evaluation conditions are described in
Section~\ref{sec:configuration}.

%% file: sections/03-experimental-setup.tex
\section{Evaluation Design}
\label{sec:experiments}
\enlargethispage{\baselineskip}

This technical report evaluates whether a unified personal health history
can support complementary forms of disease prediction. The evaluation
covers disease trajectories, risk prioritization, and cumulative outcomes,
using a common disease dictionary and three comparison conditions.

\subsection{Disease Outcomes}
\label{sec:data}

The evaluation covers 25 disease endpoints across nine clinical categories.
Each target is defined by qualifying
ICD-10 records. The outcome is the \emph{first recorded qualifying diagnosis}
after the prediction cutoff for an eligible endpoint; it does not denote
biological onset or recurrence. The full definitions and exclusions are
given in Table~\ref{tab:endpoints} in Appendix~\ref{app:endpoints}.

The broad all-malignant-neoplasms endpoint includes the four specified
cancers: lung, colorectal, breast, and prostate cancer. These overlapping
endpoints remain distinct prediction targets. For the primary B/C
evaluation, they are combined into one cancer family, yielding 21 evaluation
families. Clinical categories, prediction endpoints, and evaluation families
therefore serve different purposes.

\subsection{Prediction Tasks}
\label{sec:tasks}

\paragraph{Task A: Disease Trajectory Forecasting.}
Predict which eligible diseases will first be recorded over the next five
years and assign each predicted disease to year 1, 2, 3, 4, or 5. The output
is a list of disease--year pairs, with at most one event per disease.
Evaluation distinguishes disease identification from agreement in event
timing.

\paragraph{Task B: Disease Risk Ranking.}
Rank the five most likely eligible diseases within a specified future
window. The windows are 6 months and 1, 2, 3, 5, and 10 years. Each window
has its own ranking; the list expresses relative priority rather than
asserting that every listed disease will occur. Fewer than five entries
are required when fewer than five candidates are eligible.

\paragraph{Task C: Multi-horizon Disease Prediction.}
Predict whether each of the 25 diseases will have been recorded by each of
the same six cumulative horizons. The output is a disease-by-horizon
binary forecast. Once a disease is predicted to occur within one horizon,
it remains positive at later horizons. This task measures how well a
personal history supports disease prediction at different time scales.

\subsection{Models and Evaluation Settings}
\label{sec:configuration}

Using UK Biobank data~\cite{bycroft2018}, we compare GPT-5.5, GPT-5.6 Terra,
GPT-5.6 Sol, and GPT-6 Astra across \emph{Baseline}, \emph{Context},
and \emph{Harness}. Baseline is
the reference condition. Context uses the accumulated chronological
history for direct reasoning, and Harness extends this formulation through
the Agent Harness (Sections~\ref{sec:context} and~\ref{sec:harness}).
The aggregate comparisons do not isolate the effects of individual Skills
or other Harness components.

Task B's scored population varies by horizon because ranking evaluation
requires sufficient follow-up and at least one relevant outcome
(Appendix~\ref{app:shared}). All horizons are evaluated relative to a
case-specific prediction cutoff. An unobserved outcome with insufficient
follow-up remains unknown and is excluded from the relevant denominator,
rather than counted as negative.

The primary measures are five-year exact-endpoint F1 for A, family-level
NDCG@5 at each horizon for B, and family-level micro-F1 for C, with five-year
C performance as the primary readout. Timing agreement, precision, recall,
and other supporting measures provide complementary views of performance.
Appendix~\ref{app:metrics} gives the essential scoring definitions.
Results are descriptive point estimates. The analysis boundaries are
discussed in Section~\ref{sec:analysis-limits}.

%% file: sections/04-results.tex
\section{Results}
\label{sec:results}
\makeatletter\setlength{\@fptop}{0pt}\makeatother

\subsection{Improved Prediction across Tasks and Models}
\begin{figure}[p]
\centering
\includegraphics[width=\linewidth]{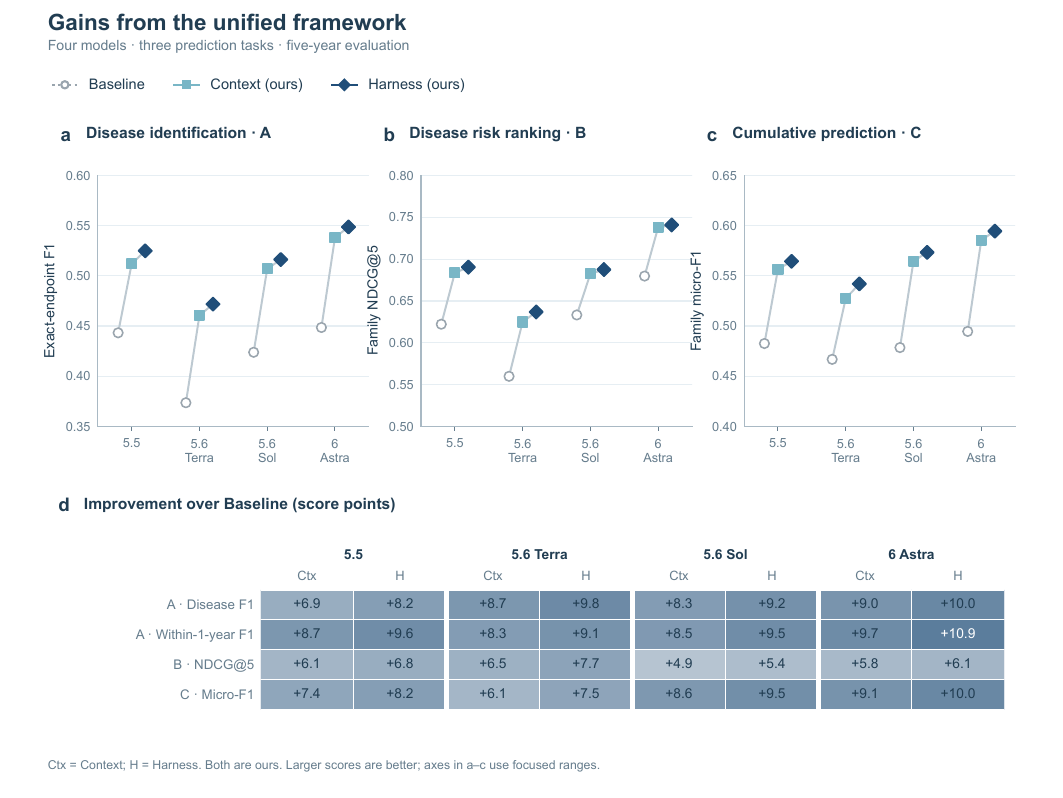}
\caption{Overall performance of Context and Harness relative to Baseline. a–c, Reported five-year primary scores for each model; task-specific focused axes retain absolute values. d, Score-point gains over Baseline for identification, year agreement, ranking and cumulative prediction. A uses exact endpoints; B/C use families. Values are point estimates; Ctx denotes Context and H denotes Harness.}
\label{fig:framework-overall}
\end{figure}

Both Context and Harness improve on Baseline in all 16 primary
model--readout comparisons (Table~\ref{tab:framework-primary};
Figure~\ref{fig:framework-overall}).
Across four models, mean disease-identification F1 in A rises from 0.422
to 0.504 with Context and 0.515 with Harness, corresponding to relative
gains of 19.4\% and 22.0\%. Five-year C micro-F1 increases from 0.481
to 0.559 and 0.569 (16.2\% and 18.3\%). Year agreement and risk ranking
also improve. Harness adds 0.30--1.45 score points over Context across
these comparisons. A score point is 100 times a raw score difference;
A uses exact endpoints and primary B/C scores use disease families.

\input{results/tables/primary_comparison}

\subsection{Performance across Prediction Horizons}
\begin{figure}[p]
\centering
\includegraphics[width=\linewidth]{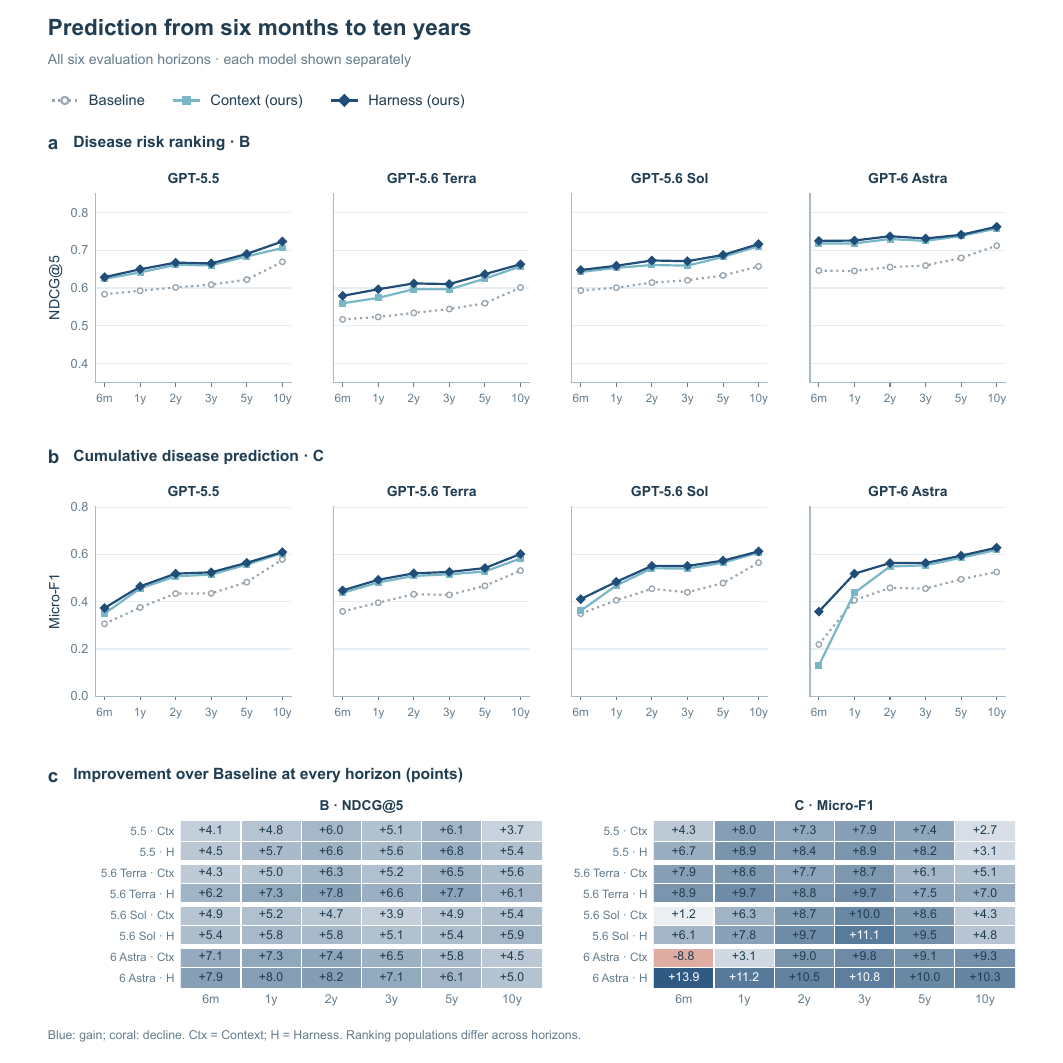}
\caption{Performance across six cumulative horizons. a–b, Three-condition curves for every model, using family NDCG@5 for B and family micro-F1 for C. Horizons are shown as equally spaced categories; y-ranges are shared within each task. c, Gains over each model's Baseline, with a common symmetric ±15-point color scale. Negative values are retained. B uses horizon-specific scored populations; curves do not track a fixed cohort.}
\label{fig:framework-horizons}
\end{figure}

The gains extend across six horizons, from six months to ten years
(Figure~\ref{fig:framework-horizons}). In B, both Context and Harness
exceed Baseline in all 24 model--horizon comparisons. In C, Harness
improves in all 24, and Context in 23; its exception is Astra at six
months, with an 8.81-point decrease. Mean Harness gains over Baseline
in C range from 6.27 to 10.14 points across horizons.
Task B uses horizon-specific scored populations, so these profiles
do not track a fixed cohort.

\subsection{Broad Gains across Disease Outcomes}
\label{sec:analysis-limits}
\begin{figure}[p]
\centering
\includegraphics[width=\linewidth]{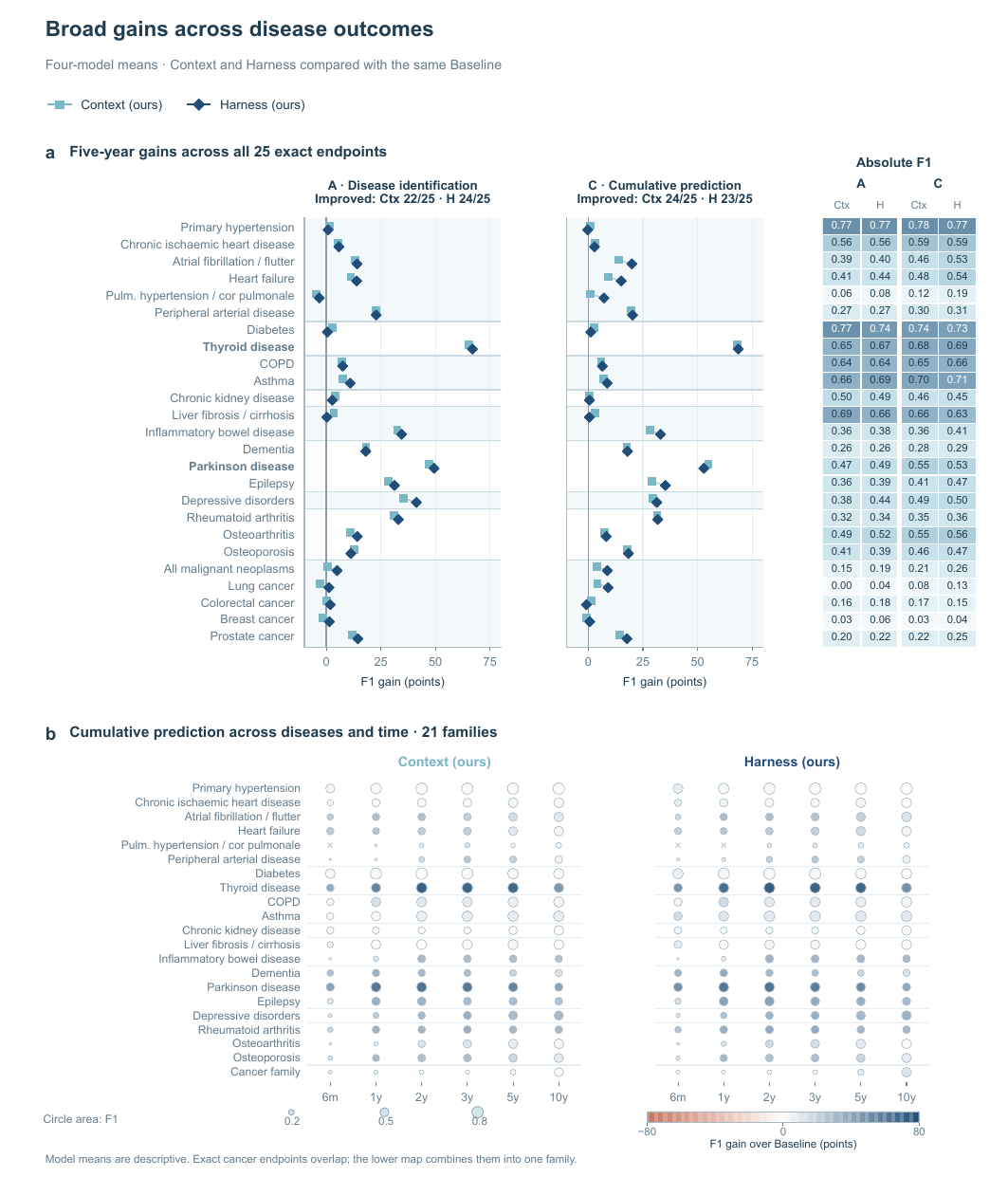}
\caption{Disease coverage and temporal variation. All values are equal-weight means of four model-specific scores, not pooled participant estimates. a, Context and Harness F1 gains over Baseline for all 25 exact endpoints at five years, with absolute scores alongside. Improvement counts use unrounded values. b, All 21 C families across six horizons: circle area is absolute F1, and color is the gain over Baseline on a shared ±80-point scale; crosses mark zero F1. The five cancer endpoints overlap in a and form one family in b. Disease supports and paired uncertainty are unavailable.}
\label{fig:framework-diseases}
\end{figure}

Gains are distributed across the disease spectrum
(Figure~\ref{fig:framework-diseases}). Using equal-weight four-model
means, Context and Harness improve F1 over Baseline for 22 and 24 of
the 25 exact endpoints in A, and for 24 and 23 in five-year C.
Thyroid disease and Parkinson's disease have the largest A gains with
Harness: 66.94 and 49.40 points, reaching F1 of 0.669 and 0.494 from
mean Baseline F1 of zero. Effects vary by disease, model and horizon.

Higher F1 also involves tradeoffs. For Astra at five years in C,
Baseline, Context and Harness precision is 0.653, 0.653 and 0.561,
respectively, while recall is 0.398, 0.531 and 0.633.
The fraction of fully observed zero-positive cases with any false
positive rises from 0.647 to 0.721 and 0.868.
All model averages are descriptive, not pooled participant estimates.
Paired uncertainty, disease support counts and component ablations
are unavailable; these comparisons do not establish statistical
significance, effects of individual Skills or clinical utility.
\clearpage
\clearpage
\clearpage
\clearpage

%% file: results/tables/primary_comparison.tex
\begingroup\fontsize{8.3}{10.6}\selectfont
\setlength{\tabcolsep}{5pt}
\begin{longtable}{llrrrr}
\caption{Five-year primary scores. Context and Harness are two formulations of our framework.}\label{tab:framework-primary}\\
\toprule
 & & \multicolumn{2}{c}{A: trajectory forecasting} & B: ranking & C: prediction\\
Model & Condition & Disease F1 & Within-1-year F1 & NDCG@5 & Micro-F1\\
\midrule
\endfirsthead
Model & Condition & Disease F1 & Within-1-year F1 & NDCG@5 & Micro-F1\\
\midrule
\endhead
GPT-5.5 & Baseline & 0.4433 & 0.3548 & 0.6222 & 0.4826 \\
 & Context (ours) & 0.5121 & 0.4416 & 0.6836 & 0.5566 \\
 & Harness (ours) & \textbf{0.5249} & \textbf{0.4504} & \textbf{0.6903} & \textbf{0.5644} \\
\addlinespace[3pt]
GPT-5.6 Terra & Baseline & 0.3737 & 0.3184 & 0.5600 & 0.4669 \\
 & Context (ours) & 0.4605 & 0.4013 & 0.6249 & 0.5275 \\
 & Harness (ours) & \textbf{0.4719} & \textbf{0.4096} & \textbf{0.6368} & \textbf{0.5420} \\
\addlinespace[3pt]
GPT-5.6 Sol & Baseline & 0.4239 & 0.3520 & 0.6333 & 0.4786 \\
 & Context (ours) & 0.5069 & 0.4373 & 0.6823 & 0.5646 \\
 & Harness (ours) & \textbf{0.5162} & \textbf{0.4473} & \textbf{0.6875} & \textbf{0.5734} \\
\addlinespace[3pt]
GPT-6 Astra & Baseline & 0.4486 & 0.3703 & 0.6797 & 0.4947 \\
 & Context (ours) & 0.5381 & 0.4669 & 0.7378 & 0.5855 \\
 & Harness (ours) & \textbf{0.5487} & \textbf{0.4798} & \textbf{0.7408} & \textbf{0.5946} \\
\addlinespace[3pt]
\bottomrule
\end{longtable}
{\footnotesize A uses exact endpoints; B/C use disease families. Bold marks the highest score within each model and readout.\par}
\endgroup

%% file: sections/05-future-work.tex
\section{Future Work}
\label{sec:future}

\subsection{Counterfactual Interventions and Personal Health Management}

The event-based representation provides a basis for studying how a person's
future health might respond to a change in behavior, environment, or treatment.
An intervention can be specified relative to the observed Lifelong Health
Context, with its timing and affected factors made explicit. Estimating an
intervention effect requires a causal estimand, identification assumptions,
and a validation design that addresses confounding, selection, and changes
in measurement intensity. An altered text input and a new forecast alone
cannot establish that effect.

Future Skills could connect causal or mechanistic methods to the Health
Reasoning Agent, returning assumptions and uncertainty alongside estimates.
Observed daily Health States, simulated events, and forecasts would remain
distinct, allowing personal health management scenarios to be evaluated
against a stable account of the recorded history.

\subsection{Longitudinal Validation and Responsible Use}

Prospective and external evaluations are needed to test how Context and
Harness perform as observations accumulate. Such studies should assess
prediction quality, sensitivity to missing and conflicting evidence,
attribution to source events, and the value of individual Skills. Component
ablations and matched input conditions are needed to distinguish additional
information from the effect of tool-assisted reasoning. Such studies would clarify
when additional life-course information improves prediction and how its
contribution changes across individuals and forecast horizons.

Clinical use requires evidence that improved forecasts benefit patients.
Prospective studies should assess this benefit alongside fairness, privacy,
resource cost, and human oversight across populations and care settings.
Versioned evidence and temporal access controls can support these studies
by preserving which information informed each prediction. Such validation
would test whether whole-person digital twins can translate accumulated
health evidence into better decisions for personalized disease prevention
throughout life.

%% file: sections/06-appendix.tex
\appendix
\section{Disease Endpoint Definitions}
\label{app:endpoints}

Table~\ref{tab:endpoints} lists the 25 targets in nine clinical categories.
Categories organize the disease dictionary; evaluation families account
for overlap between targets.

\begingroup
\fontsize{9.2}{11.2}\selectfont
\setlength{\tabcolsep}{5pt}
\renewcommand{\arraystretch}{1.07}
\begin{longtable}{@{}p{0.38in}p{3.35in}p{2.50in}@{}}
\caption{Disease endpoint dictionary: nine categories and 25 endpoints.
Ranges include their component codes unless an exclusion is stated.}
\label{tab:endpoints}\\
\toprule
ID & Endpoint & ICD-10 definition\\
\midrule
\endfirsthead
\multicolumn{3}{@{}l}{\small Table~\thetable\ (continued)}\\
\toprule
ID & Endpoint & ICD-10 definition\\
\midrule
\endhead
\midrule
\multicolumn{3}{r@{}}{\small Continued on next page}\\
\endfoot
\bottomrule
\endlastfoot
\multicolumn{3}{@{}l}{\textbf{Circulatory disorders}}\\*
D01 & Essential (primary) hypertension & I10\\
D02 & Chronic ischaemic heart disease & I25, excluding I25.2\\
D03 & Atrial fibrillation and flutter & I48\\
D04 & Heart failure & I50\\
D05 & Pulmonary hypertension / chronic cor pulmonale & I27\\
D06 & Peripheral arterial disease & I70, I73\\[2pt]
\multicolumn{3}{@{}l}{\textbf{Endocrine and metabolic disorders}}\\*
D07 & Diabetes mellitus & E10--E14\\
D08 & Thyroid disorders & E01--E07\\[2pt]
\multicolumn{3}{@{}l}{\textbf{Respiratory disorders}}\\*
D09 & Chronic obstructive pulmonary disease & J41--J44\\
D10 & Asthma & J45\\[2pt]
\multicolumn{3}{@{}l}{\textbf{Renal disorders}}\\*
D11 & Chronic kidney disease & N18\\[2pt]
\multicolumn{3}{@{}l}{\textbf{Digestive disorders}}\\*
D12 & Hepatic fibrosis / cirrhosis & K74, K70.3\\
D13 & Inflammatory bowel disease & K50--K51\\[2pt]
\multicolumn{3}{@{}l}{\textbf{Neurological and neurocognitive disorders}}\\*
D14 & Dementia & F00--F03, G30\\
D15 & Parkinson's disease & G20\\
D16 & Epilepsy & G40\\[2pt]
\multicolumn{3}{@{}l}{\textbf{Mental health disorders}}\\*
D17 & Depressive disorders & F32--F33\\[2pt]
\multicolumn{3}{@{}l}{\textbf{Musculoskeletal and immune disorders}}\\*
D18 & Rheumatoid arthritis & M05--M06\\
D19 & Osteoarthritis & M15--M19\\
D20 & Osteoporosis & M80--M81\\[2pt]
\multicolumn{3}{@{}l}{\textbf{Malignant neoplasms}}\\*
D21 & All malignant neoplasms & C00--C97\\
D22 & Lung cancer & C34\\
D23 & Colorectal cancer & C18--C20\\
D24 & Breast cancer & C50\\
D25 & Prostate cancer & C61\\
\end{longtable}
\endgroup

D22--D25 are contained within D21, whose wider definition also includes
other malignancies. Primary B/C scoring combines D21--D25 into one cancer
family and retains each of D01--D20 separately, giving 21 families.
The dictionary does not add acute myocardial infarction or stroke as
separate targets, or K72.1 as a standalone qualifying code for D12.
Diabetes includes types beyond type 2; thyroid disorders and dementia are
combined endpoints under their listed codes. Breast and prostate targets
retain sex-specific eligibility. Prevalent and other ineligible endpoints
are excluded from incident prediction denominators.

\clearpage
\begingroup
\titlespacing*{\paragraph}{0pt}{6pt}{1em}
\section{Essential Evaluation Details}
\label{app:metrics}

\subsection{Eligibility and Follow-up}
\label{app:shared}
Targets are first qualifying diagnoses recorded strictly after the
prediction cutoff; records on that date belong to prior history.
An eligible endpoint is positive if observed by the horizon, negative only
with sufficient event-free follow-up, and otherwise unknown.
Horizons use calendar months or years; follow-up accounts for death, loss
to follow-up, and record coverage. Unknown and ineligible cells are excluded
from denominators; missing predictions are not negatives.
Task B additionally requires a valid ranking, complete follow-up, and at
least one eligible candidate and positive family, yielding horizon-specific
scored populations.

\subsection{Scoring Definitions}
\label{app:task-a}
\paragraph{Task A.}
Disease-set F1 ignores year and uses exact endpoints. Strict temporal
matching requires the same endpoint and first-event year; tolerant matching
allows $|\widehat{j}-j|\leq1$. Unobserved follow-up intervals are excluded.
A predicted specific cancer derives D21 in the same year only if D21 is
eligible and has not occurred or been predicted earlier. Explicitly
redundant same-year parent--child pairs are not permitted.

\paragraph{Task B.}
\label{app:task-b}
Let $K_i=\min(5,M_i)$ for $M_i$ eligible candidates, and $R_i>0$ be the
number of true positive families. For an endpoint at rank $r$, $z_r=1$
only if its family is positive and has not appeared earlier.
Duplicate families retain their original ranks without further credit:
\begin{equation}
 \mathrm{NDCG@5}_i=
 \frac{\sum_{r=1}^{K_i} z_r/\log_2(r+1)}
 {\sum_{r=1}^{\min(K_i,R_i)}1/\log_2(r+1)}.
 \label{eq:ndcg}
\end{equation}
Recall@5 is $\sum_r z_r/R_i$; MRR averages the reciprocal of the first
credited positive rank, or zero if none is retrieved. Ranking metrics
are averaged over scored cases. The exact view omits family collapse.
Disease contributions sum to aggregate NDCG within the same scoring view.

\paragraph{Task C.}
\label{app:task-c}
An eligible D21 must be positive when an eligible specific cancer is
predicted positive; ineligible D21 stays zero and is excluded.
A family's prediction is the logical OR over eligible members.
Its truth is positive if any member is observed positive, negative only
if all are evaluably negative, and otherwise unknown.
Each horizon is scored separately; observed positive cells remain
evaluable when other cells are unknown.

\paragraph{Pooled and subgroup metrics.}
\label{app:aggregation}
For true positives (TP), false positives (FP), and false negatives (FN),
micro metrics pool eligible, evaluable counts before division:
\begin{equation}
 P=\frac{\mathrm{TP}}{\mathrm{TP}+\mathrm{FP}},\quad
 R=\frac{\mathrm{TP}}{\mathrm{TP}+\mathrm{FN}},\quad
 F_1=\frac{2\mathrm{TP}}{2\mathrm{TP}+\mathrm{FP}+\mathrm{FN}}.
 \label{eq:prf}
\end{equation}
Primary A uses exact endpoints; primary C uses families.
Fully observed zero-positive cases contribute to micro false positives;
their any-false-positive fraction and mean predicted-positive family count
use that subgroup's denominator. Analysis limits are given in
Section~\ref{sec:analysis-limits}.

\paragraph{Random score-scale reference.}
\label{app:random-reference}
The graphical abstract uses an independent, outcome-enriched label
reference sample, separate from model evaluation.
For independent positive guesses with probability $q=0.5$ and evaluable
family-cell prevalence $p_h$, $F_{1,h}\approx2p_hq/(p_h+q)$ in large samples.
At 6, 12, 24, 36, 60, and 120 months, the reference F1 is
0.137, 0.183, 0.221, 0.246, 0.279, and 0.343.
These analytical values are score-scale references, not a random-prediction
experiment, a universal F1 threshold, or a matched-cohort comparison.
\par
\endgroup